\documentclass[11pt,a4paper]{article}
\usepackage[hyperref]{eacl2021}
\usepackage{times}
\usepackage{latexsym}
\usepackage{graphicx}

\usepackage{microtype}

\aclfinalcopy 

\title{Towards Zero-shot Cross-lingual Image Retrieval}

\author{Pranav Aggarwal \\
  Adobe Inc. \\
  San Jose, CA \\
  \texttt{pranagga@adobe.com} \\\And
  Ajinkya Kale \\
  Adobe Inc. \\
  San Jose, CA \\
  \texttt{akale@adobe.com} \\}

\date{}

\begin{document}
\maketitle

\begin{abstract}

There has been a recent spike in interest in multi-modal Language and Vision problems. On the language side, most of these models primarily focus on English since most multi-modal datasets are monolingual. We try to bridge this gap with a zero-shot approach for learning multi-modal representations using cross-lingual pre-training on the text side. We present a simple yet practical approach for building a cross-lingual image retrieval model which trains on a monolingual training dataset but can be used in a zero-shot cross-lingual fashion during inference. We also introduce a new objective function which tightens the text embedding clusters by pushing dissimilar texts from each other. Finally, we introduce a new 1K multi-lingual MSCOCO2014 caption test dataset (XTD10) in 7 languages that we collected using a crowdsourcing platform. We use this as the test set for evaluating zero-shot model performance across languages. XTD10 dataset is made publicly available here: \url{https://github.com/adobe-research/Cross-lingual-Test-Dataset-XTD10}

\end{abstract}

\section{Introduction}
\begin{figure}[t]
\begin{center}
   \includegraphics[width=1\linewidth]{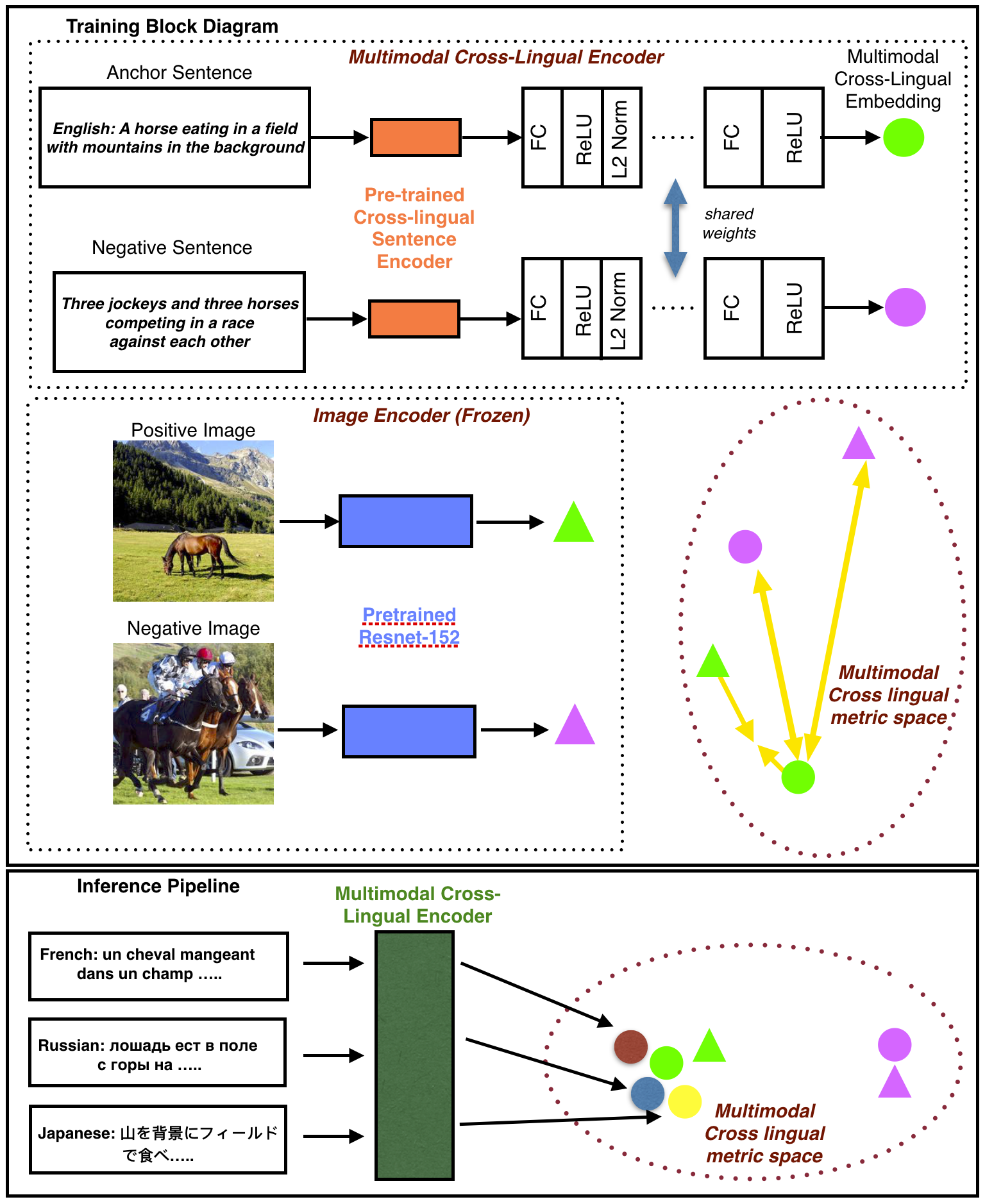}
\end{center}
   \caption{Overview of our approach.}
\label{fig:overview}
\end{figure}

Image retrieval is a well studied problem in both academia and industry ~\cite{datta2008image, jing2015visual, yang2017visual, zhang2018visual, shankar2017deep}. Most research looks at image retrieval in a monolingual setup for a couple of reasons:

\begin{itemize}
    \item Lack of multi-lingual Vision-Language datasets supporting a wide range of languages
    \item Extensibility towards new and low-resource language support  
\end{itemize}

Multi-lingual dataset collection has always been a major hurdle when it comes to building models in a one-model-fits-all style that can provide good results for image retrieval across multiple languages. Most methods ~\cite{rotman-etal-2018-bridging, DBLP:journals/corr/NakayamaN16, DBLP:journals/corr/abs-1809-07615} rely on direct translations of English captions while others ~\cite{gella-etal-2017-image, Huang2019MultiHeadAW} have used independent image and language text pairs. Based on previous research learning we try to explore following ideas in this paper:

\begin{itemize}
    \item \textit{One-model-fits-all}: Can we use pre-trained cross-lingual embeddings with monolingual image-text training data to learn representations in a common embedding space for image retrieval?
    \item \textit{Multi-lingual Eval Dataset}: Build an evaluation set for multi-lingual image retrieval to test in a zero-shot retrieval setup.
\end{itemize}

In our approach we try to take advantage of the recent developments in cross-lingual sentence embeddings ~\cite{DBLP:journals/corr/abs-1812-10464, DBLP:journals/corr/abs-1907-04307} which are effective in aligning multiple languages in a common embedding space. Due to scarcity of multi-lingual sentence test datasets, for evaluation we combine 10 non-English language annotations to create a Cross-lingual Test Dataset called \href{https://github.com/adobe-research/Cross-lingual-Test-Dataset-XTD10}{XTD10}.

\section{Related Work}
In an image retrieval system, image metadata (like tags) is often noisy and incomplete. As a result, matching low-level visual information with the user's text query intent have become popular over time for large scale image retrieval tasks~\cite{10.1145/1282280.1282283}. Metric learning is one way to achieve this, by projecting samples from different modalities in a common semantic representation space.
~\cite{CCA} is an early mono-lingual method that does this.

One of the primary goals of our exploration is multi-lingual retrieval support and hence multi-lingual multi-modal common representations become a key aspect of our solution. 
Recent multi-lingual metric learning methods ~\cite{calixto-liu-2017-sentence, DBLP:journals/corr/abs-1809-07615} have tried to minimize distance between image and caption pairs as well as multi-lingual pairs of text. These approaches are limited based on availability of large parallel language corpora. ~\cite{gella-etal-2017-image} uses images as pivots to perform metric learning which allows text in one language to be independent of the other languages. ~\cite{Huang2019MultiHeadAW} uses a similar approach but adds visual object detection and multi-head attention~\cite{DBLP:journals/corr/VaswaniSPUJGKP17} to selectively align salient visual objects with textual phrases to get better performance. Similarly ~\cite{Kim2020MULEMU, Mohammadshahi2019AligningMW} use language-independent text encoders to align different languages to a common sentence embedding space using shared weights along with simultaneously creating a multi-modal embedding space. This approach allows better generalization. We use ~\cite{cross-modal-learning} as our baseline as this was the only method we found which followed a zero shot approach, but on word level.

Datasets like ~\cite{yoshikawa-etal-2017-stair, COCO-CN, elliott-etal-2016-multi30k, Grubinger06theiapr, AIC-ICC} provide data with images for only 2-3 languages which do not help in scaling models to a large number of languages. Due to this, most of the models discussed above work with limited language support. Work that comes closest to ours is the Massively Multilingual Image Dataset ~\cite{hewitt-etal-2018-learning} initiative. A big difference being they focus on parallel data for 100 languages, but the dataset is word-level concepts losing out on inter-concept and inter-object context within complex real world scenes that captions provide.


\section{Proposed Method}
Towards solving the issues discussed above for multi-lingual image retrieval, we take a simple yet practical and effective zero-shot approach by training the model with only English language text-image pairs using metric learning to map the text and images in the same embedding space. We convert the English text training data into its cross-lingual embedding space for initialization which helps support multiple languages during inference. Figure \ref{fig:overview} gives an overview of our approach.

\subsection{Model Architecture}
Most industry use cases involving images will have a pre-trained image embeddings model. It is expensive to build and index new embeddings per use case. Towards this optimization and without loss of generality, we assume there exists a pre-trained image embedding model like ResNet~\cite{DBLP:journals/corr/HeZRS15} trained on ImageNet~\cite{ImageNet}. We keep the image embedding extraction model frozen and do not add any trainable layers on the visual side. From pre-trained ResNet152 architecture, we use the last average pooled layer as the image embedding of size 2048. 

On the text encoder end, we first extract the sentence level embeddings for the text data. We experiment with two state-of-the-art cross-lingual models - LASER ~\cite{DBLP:journals/corr/abs-1812-10464} and Multi-lingual USE (or mUSE) ~\cite{DBLP:journals/corr/abs-1907-04307, DBLP:journals/corr/abs-1810-12836}. LASER uses a language agnostic BiLSTM~\cite{LSTM} encoder to create the sentence embeddings and supports sentence-level embeddings for 93 languages.
mUSE is a Transformer \cite{DBLP:journals/corr/VaswaniSPUJGKP17} based encoder which supports sentence-level embeddings for 16 languages. It uses a shared multi-task dual-encoder training framework for several downstream tasks 

After the sentence-level embedding extraction, we attach blocks of fully connected layer with dropout~\cite{dropout}, rectified linear units (ReLU) activation layer~\cite{pmlr-v15-glorot11a} and l2-norm layer, in that order. The l2-norm layer helps to keep the intermediate feature values small. For the last block, we do not add l2-norm layer to be consistent with ResNet output features. From our experiments, 3 stacked blocks gave us the best results.

\subsection{Training Strategy}
For each text caption (anchor text) and (positive) image pair, we mine a hard negative sample within a training mini-batch using the online negative sampling strategy from ~\cite{DBLP:journals/corr/abs-1905-13339}. We treat the caption corresponding to the negative image as the hard negative text. 

We propose a new objective loss function called ``Multi-modal Metric Loss (M3L)'' which helps to reduce the distance between anchor text and its positive image, while pushing away negative image and negative text from the anchor text. 
\begin{equation}
             L_{M3} = \frac{\alpha_1 * d(te_{an}, im_p)^\rho}{d(te_{an}, im_n)^\rho} + \frac{\alpha_2 * d(te_{an}, im_p)^\rho}{d(te_{an}, te_n)^\rho}
    \end{equation}
Here $te_{an}$ is the text anchor, $te_n$ is the negative text, while $im_p$, $im_n$ are the positive and negative images, respectively. d(x,y) is the square distance between x and y. $\rho$ controls the sensitivity of the change of distances. $\alpha_1$ and $\alpha_2$ are the scaling factors for each negative distance modality. For our experiments we see that when  $\rho$ = 4, $\alpha_1$ = 0.5 and $\alpha_2$ = 1, we get the best results.
To confirm its efficiency, we compare our results with another metric learning loss called ``Positive Aware Triplet Ranking Loss (PATR)'' ~\cite{DBLP:journals/corr/abs-1905-13339} which performs a similar task without negative text. 
\begin{equation}
             L_{PATR} = d(te_{an}, im_p) + max(0, \eta - d(te_{an}, im_n))
    \end{equation}

here $\eta$ penalizes the distance between anchor and negative image, therefore controlling the tightness of the clusters. In our experiments $\eta$ = 1100 gave the best performance.

We use a learning rate of 0.001 along with Adam Optimizer ~\cite{Kingma2015AdamAM} (beta1=0.99). We add a dropout of [0.2, 0.1, 0.0] for each of the fully connected layers of dimension [1024, 2048, 2048], respectively. As we want hard negatives from our mini-batch, we take a large batch size of 128 and train our model for 50 epochs. 

\section{Evaluation}
\subsection{Dataset}
We have used MSCOCO 2014 ~\cite{DBLP:journals/corr/LinMBHPRDZ14} along with our human annotated dataset XTD10 (Table \ref{fig:language}) for testing and Multi30K dataset ~\cite{elliott-EtAl:2017:WMT} for our experiments.
\subsubsection{MSCOCO 2014 and XTD10}
For MSCOCO2014 dataset, we use the train-val-test split provided in ~\cite{rajendran-etal-2016-bridge}. To convert the test set into 1K image-text pairs, for each image we sample the longest caption. As MSCOCO2014 dataset is only in English, to evaluate our model we use the French and German translations provided by ~\cite{rajendran-etal-2016-bridge}, Japanese annotations for the 1K images provided by ~\cite{yoshikawa-etal-2017-stair} and for the remaining 7 languages, we got 1K test human translated captions \footnote{we used https://www.lionbridge.com/}. Except for Japanese, all other languages are direct translations of the English test set. Together, we call this test set the Cross-lingual Test Dataset 10 (XTD10).

\subsubsection{Multi30K}
In Multi30K dataset, every image has a caption in English, French and German. We split the data 29000/1014/1000 as train/dev/test sets.
\begin{table}
\centering
\caption{Test Dataset languages and families}
\begin{tabular}{ |l|l| } 
 \hline
 Language & Family\\
 \hline
 English(en) German(de) & Germanic \\
\hline
French(fr) Italian (it) & Latin\\
 Spanish (es)  &  \\ \cline{2-2}
\hline
Korean(ko)  & Koreanic\\
\hline
Russian(ru) Polish(pl)   & Slavic\\
\hline
Turkish(tr)   & Turkic\\
\hline
Chinese Simplified(zh)   & Sino-Tibetan\\
\hline
Japanese(ja)   & Japonic\\
\hline
\end{tabular}
\label{fig:language}
\end{table}

\subsection{Quantitative Results}
\begin{table*}[t]
\centering
\caption{Image Retrieval Recall @10 on the XTD10 dataset for 11 different languages including English}
\begin{tabular}{ | l| c| c| c| c| c| c| c| c| c| c| c| c| c| c| c| c| c| c| c| c| c| c|} 
 \hline
 Model & en & de & fr & it & es & ru & ja & zh & pl & tr & ko\\
 \hline
 $LASER_{PATR}$& 0.803 & 0.702 & 0.686 & 0.673 & 0.682 & 0.677 & 0.572 & 0.672 & 0.666 & 0.597 & 0.518\\
\hline
 $mUSE_{PATR}$& 0.836 & 0.712 & 0.756 & 0.769 & 0.761 & 0.734 & 0.643 & 0.736 & \textbf{0.718} & 0.669 & 0.694\\
 \hline
 $LASER_{M3L}$& 0.815 & 0.706 & 0.712 & 0.701 & 0.714 & 0.686 & 0.583 & 0.717 & 0.689 & 0.652 & 0.533\\
\hline
 $mUSE_{M3L}$& \textbf{0.853} & \textbf{0.735} & \textbf{0.789} & \textbf{0.789} & \textbf{0.767} & \textbf{0.736} & \textbf{0.678} & \textbf{0.761} & 0.717 & \textbf{0.709} & \textbf{0.707} \\
\hline
\end{tabular}
\label{fig:mscoco_quant}
\end{table*}


\begin{table}
\caption{Image Retrieval Recall @10 Multi30K dataset}
\begin{tabular}{ | l| c| c| c| c|} 
 \hline
 Model & train langs & en & de & fr \\
 \hline
 Baseline& en,fr,de & 0.546 & 0.450 & 0.469\\
\hline
 Ours & en,fr,de &  0.581 & \textbf{0.522} & \textbf{0.572}\\
 \hline
 Baseline& en & 0.566 & 0.442 & 0.460\\
\hline
 Ours & en & \textbf{0.591} & 0.488 & 0.548\\
 
\hline
\end{tabular}
\label{fig:multi30k_quant}
\end{table}

\begin{figure*}[t]
\begin{center}
   \includegraphics[width=1.06\linewidth]{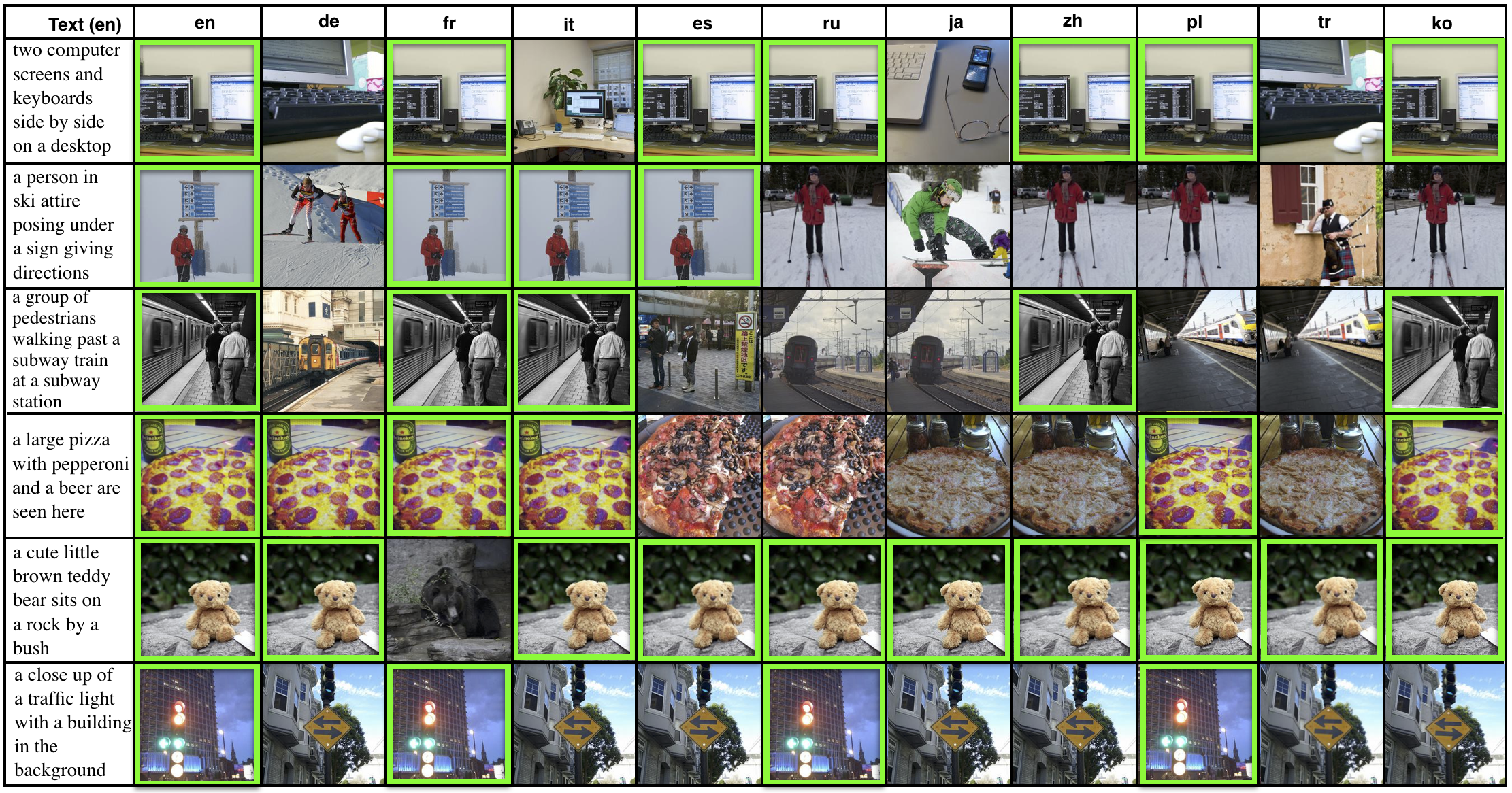}
\end{center}
   \caption{Recall@1 qualitative results. We have mentioned only English language captions in this figure. The results which are correctly ranked as 1 for their corresponding language's caption are bordered green. }
\label{fig:R@1}
\end{figure*}

\begin{figure}[t!]
\begin{center}
  \includegraphics[width=1\linewidth]{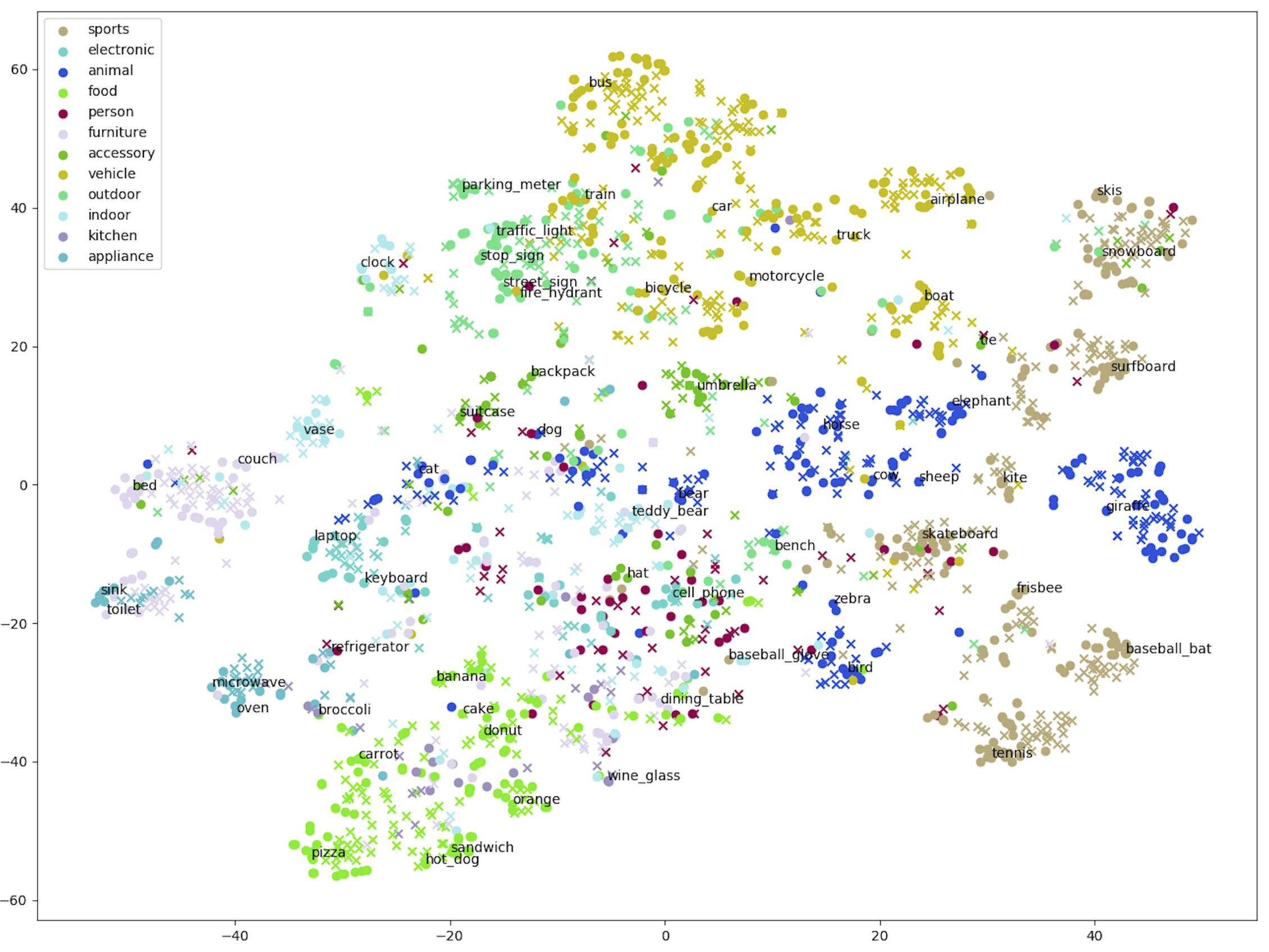}
\end{center}
  \caption{Visualization of the Multi-modal Cross-lingual space: T-SNE Scatter Plot for Italian text embeddings("o") overlayed on ResNet152 Image embeddings("x"). Subcategories are plotted based on ResNet152 clustering (best seen in color).}
\label{fig:resnet_scatter}
\end{figure}

\begin{figure}[t!]
\begin{center}
  \includegraphics[width=1\linewidth]{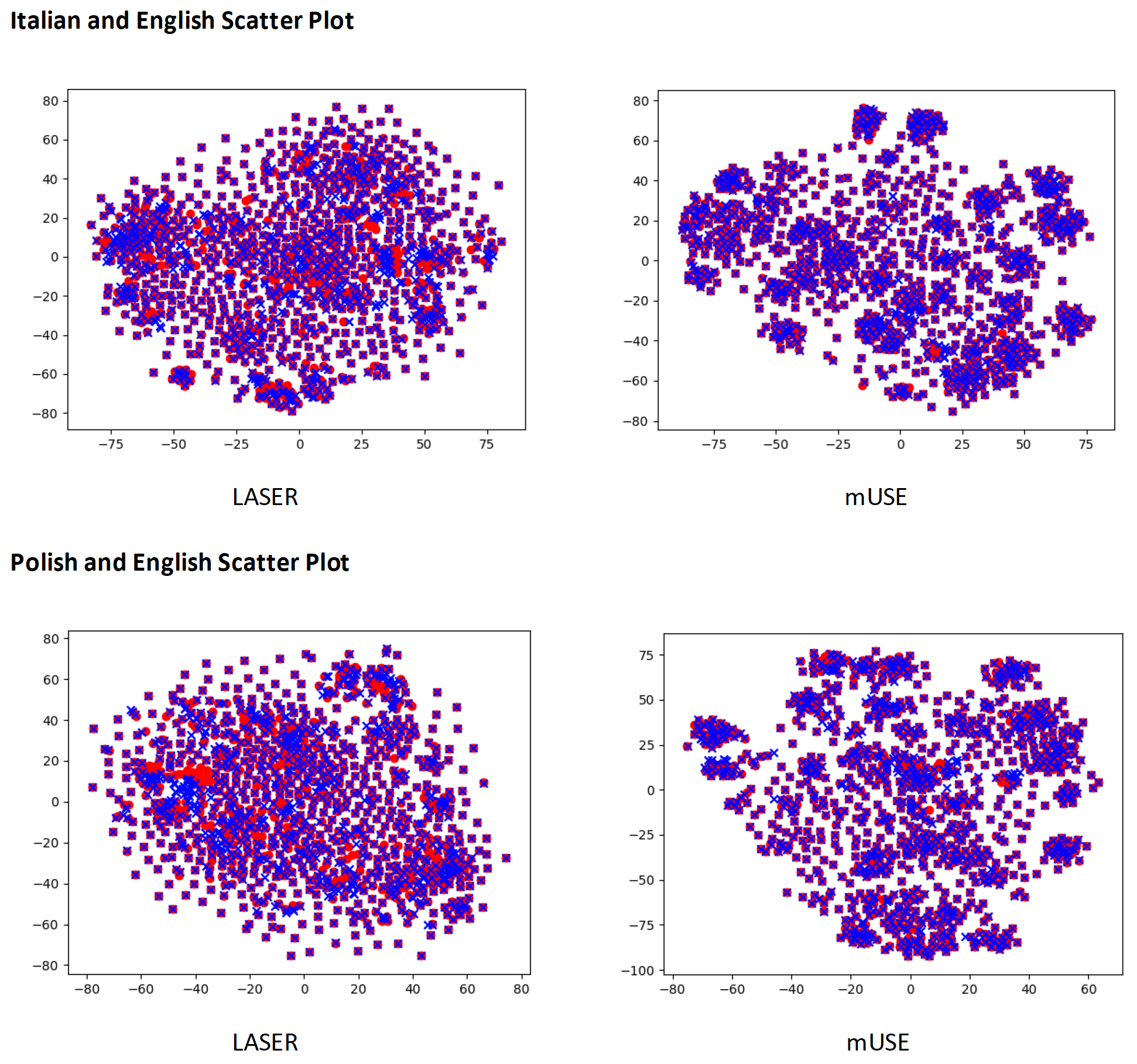}
\end{center}
  \caption{Visualization of the Cross-lingual space: Multi-lingual USE vs LASER: T-SNE Scatter Plots represent non-English cross-lingual embeddings ("x") overlayed on their respective English cross-lingual embeddings ("o").}
\label{fig:non-english-scatter}
\end{figure}

We report the Recall@10 using XTD10 dataset for each of the models trained only on English language in Table \ref{fig:mscoco_quant}. We show the comparison between LASER and mUSE sentence-embeddings using PATR and M3L metric learning losses. Because the train data is only in English, we see stronger performance with English which acts as our upper bound. With zero-shot learning, we obtain comparable performance for all the other 10 non-English languages that we test on. We observe best results when applying M3L + mUSE as the addition of negative text in the loss function creates tighter text clusters in the metric space. However, for Slavic languages like Russian and Polish, we don’t see much difference in performance when using mUSE with negative text in English.
On clustering the cross-lingual non-English embeddings with respective English embeddings for both mUSE and LASER, we saw very clear overlapping for mUSE compared to LASER across all languages (please refer supplementary material for plots). Therefore, we see consistency in the performance across all languages for models trained with mUSE. We also see a similar trend in performances reported in Table \ref{fig:mscoco_quant} between mUSE and LASER in Table 7 in ~\cite{DBLP:journals/corr/abs-1907-04307}. We also suspect that because LASER is trained with more languages and is a less complex model, it generalizes more than mUSE. 

We also report the Recall@10 using Multi30K dataset with our model trained on M3L + mUSE and ~\cite{cross-modal-learning} as our baseline in Table \ref{fig:multi30k_quant}. The Baseline uses MUSE ~\cite{lample2017unsupervised} cross-lingual word embeddings for training. Our method outperforms Baseline in both multi-lingual and zero-shot training. Comparing multi-lingual vs zero-shot training, we see that training with all languages decreases performance for English due to model generalization. In zero-shot setup, the recall accuracies for German and French dip slightly yet are comparable with the multi-lingual training result.

\subsection{Qualitative Results}
In Figure \ref{fig:non-english-scatter} we see that the English and non-English mUSE cross-lingual embeddings are more tightly clustered with each other as compared to LASER embeddings. This explains why we get retrieval results for non-English languages closer to its English counterpart for mUSE as compared to LASER. We use $USE_{M3L}$ model results trained on MSCOCO2014 dataset for next two observations. Figure \ref{fig:resnet_scatter} demonstrates the visual embeddings alignment with Italian multi-modal cross-lingual clustering. In Figure \ref{fig:R@1} we have visualized the images retrieved at Rank 1 for 11 languages. Our model is able to capture all the objects in most results across all languages. The R@1 for $3^{rd}$ and $6^{th}$ example for languages which do not give the desired image as Rank 1, still cover all the object concepts in their corresponding text captions. For the $5^{th}$ example, we see that for French, the image at R@1, the 2 objects ``rock'' and ``bush'' are covered in the image, but not ``teddy bear''. This is because its French caption ``\textit{un petit ourson brun mignon assis sur un rocher par un buisson}'' doesn't cover ``teddy'' concept as when translated to English it says "a cute little brown bear sitting on a rock by a bush". But we do get the desired result when we use the Google translated~\footnote{https://translate.google.com/} English to French caption instead.

\section{Conclusion}
We propose a zero-shot setup for cross-lingual image retrieval and evaluated our models with 10 non-English languages. This practical approach can help scale to languages in scenarios where multi-lingual training data is scarce. In future we plan to investigate few-shot setup where some training data is available per language and fine-tuning the image side along with the text side in an end-to-end fashion further improves the retrieval accuracy. 

\section{Acknowledgment}
We thank Tracy King for guiding us through this project and Mayank Dutt for assisting in the dataset annotation process.

\bibliography{anthology,eacl2021}
\bibliographystyle{acl_natbib}

\end{document}